\documentclass[runningheads]{llncs}

\usepackage[normalem]{ulem}
\usepackage{graphicx}  
\usepackage{multirow}  
\usepackage{tabularx}  
\usepackage{array}     
\usepackage{caption}   
\usepackage{amsmath}   
\usepackage{booktabs}  
\usepackage{float}     
\usepackage{graphics}
\usepackage{graphicx}
\usepackage{amsmath} 
\usepackage{listings}
\usepackage{array}
\usepackage{booktabs}
\usepackage{subcaption}

\usepackage[margin=3.50cm]{geometry}
\begin{document}
\linespread{0.9}

\newcommand{\Baseline}{baseline}
\newcommand{\Surrogate}{surrogate}
\newcommand{\Expensive}{expensive}

\newcommand{\BaselineCap}{Baseline}
\newcommand{\SurrogateCap}{Surrogate}
\newcommand{\ExpensiveCap}{Expensive}

\title{NeuroLGP-SM: A Surrogate-assisted Neuroevolution Approach using Linear Genetic Programming}

\author{Fergal Stapleton\inst{1}\orcidID{0000-0002-5347-1573}, Brendan Cody-Kenny\inst{2} \and \\
Edgar Galv\'an\inst{1}\orcidID{0000-0001-8474-5234} }

\authorrunning{F. Stapleton, Brendan Cody-Kenny and E. Galv\'an}
%
\institute{Naturally Inspired Comp. Res. Group, CS Department, Hamilton Institute, Maynooth University,  IE \\
\email{fergal.stapleton.2020@mumail.ie, edgar.galvan@mu.ie} \and
Sema, Baltimore, MD, USA \\
\email{brendan@semasoftware.com}}






\titlerunning{NeuroLGP-SM}
\maketitle

\begin{abstract}
Evolutionary algorithms are increasingly recognised as a viable computational approach for the automated optimisation of deep neural networks (DNNs) within artificial intelligence. This method extends to the training of DNNs, an approach known as neuroevolution. However, neuroevolution is an inherently resource-intensive process, with certain studies reporting the consumption of thousands of GPU days for refining and training a single DNN network. To address the computational challenges associated with neuroevolution while still attaining good DNN accuracy, surrogate models emerge as a pragmatic solution. Despite their potential, the integration of surrogate models into neuroevolution is still in its early stages, hindered by factors such as the effective use of high-dimensional data and the representation employed in neuroevolution. In this context, we address these challenges by employing a suitable representation based on Linear Genetic Programming, denoted as NeuroLGP, and leveraging Kriging Partial Least Squares. The amalgamation of these two techniques culminates in our proposed methodology known as the NeuroLGP-Surrogate Model (NeuroLGP-SM). For comparison purposes, we also code and use a baseline approach incorporating a repair mechanism, a common practice in neuroevolution. Notably, the baseline approach surpasses the renowned VGG-16 model in accuracy. Given the computational intensity inherent in DNN operations, a singular run is typically the norm. To evaluate the efficacy of our proposed approach, we conducted 96 independent runs spanning a duration of 4 weeks. Significantly, our methodologies consistently outperform the baseline, with the SM model demonstrating superior accuracy or comparable results to the NeuroLGP approach. Noteworthy is the additional advantage that the SM approach exhibits a 25$\%$ reduction in computational requirements, further emphasising its efficiency for neuroevolution.


\end{abstract}

\section{Introduction}



In the preceding decades, Evolutionary Computation (EC)~\cite{EibenBook2003} has attracted considerable attention in the burgeoning field of Neural Architecture Search (NAS)~\cite{Galvn2021NeuroevolutionID}. The combination of these two fields, commonly referred to as neuroevolution, is centred around automatically finding architectures of artificial neural networks (ANNs) by evolving network topologies, hyperparameters and/or weights. Typically, the architectures are evolved based on accuracy, but often consider other design characteristics, such as network latency~\cite{chen2023run}. Neuroevolution of deep neural networks (DNNs)~\cite{lecun2015deeplearning}, has seen growing interest in recent years as discussed in a recent survey~\cite{Galvn2021NeuroevolutionID}, reviewing over 170 works, and has applications to state-of-the-art and emerging technologies such as in the field of autonomous vehicles~\cite{galvan2023evolutionary,stapleton2022neuroevolutionary}.

When considering DNNs, in general, whether using random search~\cite{bergstra2012random}, neuroevolution~\cite{Galvn2021NeuroevolutionID} or other approaches, there is a high computational overhead required in order to effectively search for well-performing architectures. To tackle this, research has turned to the use of surrogate-assisted evolutionary algorithms (SAEAs)~\cite{JIN201161}. SAEAs can be used to estimate the fitness of expensive DNNs without requiring them to be fully trained. To build an effective modelling strategy for fitness estimation, we need to impute the accuracy of untrained or partially trained networks based on knowledge from previously fully-trained neural networks. However, comparisons between models often require ingenious encoding strategies~\cite{8744404} to compare genotypes as generally, it is not possible to create distance metrics for differing network topologies~\cite{pheno_dist_kernel}. A more natural approach is to consider the behaviour of the DNN architectures rather than relying on genotypic information~\cite{hagg2019prediction,pheno_dist_kernel}, however, to date, SAEAs based around phenotypic distance metrics have focused primarily on ANNs of only one to two layers of fully-connected layers~\cite{pheno_dist_kernel}. Another limitation of this previous work is that Kriging, which is an interpolation technique widely used in SAEAs and is well suited for optimization problems~\cite{bhosekar2018advances}, suffers computational overhead when using high-dimensional data which puts further restrictions on the size of phenotypic distance vectors that can be used. 



The primary objective of this paper is to provide insight into the effective integration of surrogate models into neuroevolution, addressing a significant challenge posed by the commonly encountered high-dimensional data. The key contributions of this work are:
\begin{itemize}
\item Firstly, we implement and validate a baseline model employing a repair mechanism, a strategy frequently utilised in neuroevolutionary methods. This model surpasses the well-established VGG-16 model, setting a high-performance benchmark for comparison with our proposed approaches.

\item Secondly, we introduce an innovative representation based on Linear Genetic Programming~\cite{brameier2007linear}, termed NeuroLGP, facilitating the automatic discovery of well-performing DNNs that outperform the baseline model. The NeuroLGP approach is employed to compute the fitness of the entire population and serves as an excellent method to compare against a surrogate model (SM).

\item Thirdly, to address the challenge of high-dimensional data and enable the use of a SM to reduce computational demands in neuroevolution, we employ Kriging Partial Least Squares~\cite{bouhlel2016improving}. The fusion of this technique with NeuroLGP results in our second proposed approach, referred to as NeuroLGP-SM. This approach consistently identifies DNNs that perform similarly to those discovered by NeuroLGP, simultaneously reducing fitness evaluations and training time required for these DNNs. In our previous work~\cite{2023_GECCO_LBA_KPLS_FergalStapleton}, we identified that it was not possible to use the original Kriging approach to this end.


\item Fourthly, we demonstrate the effectiveness of our approach through a robust comparison involving 8 independent runs for each of the 2 aforementioned NeuroLGP approaches plus the baseline model, using 4 challenging image classification datasets. This extensive evaluation, deviates from the norm of a single run in the context of DNNs.

\item Our fifth key contribution lies in the innovative management of the SM. This approach remains invariant to varying network topologies and robust to data augmentation techniques. Consequently, we can train our networks with a significantly reduced number of instances while maintaining the ability to generalise effectively to unseen data.

\end{itemize}

Section~\ref{sec::related_work} details related work, Section~\ref{sec::background} discusses background. Section~\ref{sec::methodology} describes the \mbox{NeuroLGP} method along with the surrogate model in detail, Section~\ref{sec::exp} details the experimental setup, Section~\ref{sec::results} offers analysis of our results and Section~\ref{sec::conclusion} offers concluding remarks.

\section{Related Work}
\label{sec::related_work}

\subsection{Neuroevolution using Surrogate Models}



An old, but still highly relevant survey on surrogate models by Jin~\cite{JIN201161}, covers some of the fundamental aspects of surrogate-assisted EAs, such as surrogate model management strategies and acquisition functions. A more recent survey by Khaldi and Draa~\cite{khaldi2023surrogate}, covers more state-of-the-art concepts, such as considering the computational complexity of a surrogate model when performing neuroevolution. Next, we will cover some of the most relevant works.

Performance prediction can be categorized into two branches: (i) approaches that infer the performance of unseen networks based on specific characteristics of previously evaluated networks and which typically have been fully trained, and (ii) approaches that instead used partially trained information to predict the future performance of a network.

A major work in recent years of the former approach is the End-to-End Performance Predictor (E2EPP)~\cite{8744404} proposed by Sun et al. This approach uses an offline surrogate model based on random forests to search for optimal Convolutional Neural Networks (CNN) architectures. The CNN is cleverly encoded such that it maps to a numerical decision variable, which can then be processed by a random forest surrogate. This approach alleviates restrictions found in other approaches such as assuming a smooth learning curve and not requiring large amounts of training. The authors approach was shown to speed up fitness evaluations while also achieving the best performance compared to other peer performance predictors.

Approaches falling into the second category include the Freeze-Thaw Bayesian Optimization (FBO) technique proposed by Swersky et al.~\cite{swersky2014freeze}. It uses Bayesian optimization to decide if a partially trained network should be trained to completion. The fundamental idea is that a human expert is quickly able to assess if a network is likely to result in poor performance and as such can decide to halt training. Based on this notion the authors designed an approach that at its core is a form of performance prediction, where Bayesian Optimization is used to determine whether a particular partially trained network will yield preferable results compared to other networks. Unlike E2EPP, this approach does require a smooth learning curve which can be hampered if different learning rates are considered. Of interest, is that the FBO approach relies on the phenotypic behaviour of the network in order to decide which networks to evaluate.

Gaier et al.~\cite{gaier2018data} devised a kernel-based surrogate model to use with NEAT~\cite{10.1162/106365602320169811} referred to as Surrogate-assisted Neat. To overcome the limitation of variable length genotypes, they exploited a distance metric inherent in the NEAT algorithm. This distance metric which is known as the `compatibility distance' was originally designed to help promote diversity amongst the network topologies through speciation. Within this work they also use it as a distance metric to use for Gaussian Processes. They demonstrated that they were able to achieve similar performance to NEAT with much fewer evaluations.

Greenwood and McDonnell~\cite{greenwood2022surrogate} proposed a grammar-based approach which generates a tensor representation of variable length DNN topologies. An advantage of using formal grammar as a representation is that they can be designed to ensure validity of models by describing the space of allowable topologies. Their approach modifies the DeepNeat algorithm first proposed by Miikkulainen \cite{miikkulainen2019evolving}. The modelling strategy of this work is designed around a two-phase approach. Firstly, during the initialization phase, the DeepNeat algorithm is used to evolve the population of neural networks and these models are used to formulate the surrogate. Secondly, the {active learning phase} is implemented where new networks are evaluated using the surrogate model, along with a subset of networks that are fully trained and evaluated to further inform and improve the surrogate model. They noted a five times improvement in compute time. 

While NeuroLGP and NeurpLGP-SM are presented here for the first time, an analysis of the architectures discovered by these two approaches has not been presented in this work but has subsequently been addressed in our upcoming paper~\cite{Stap2406:NeuroLGP}. Additionally, this work includes additional analysis of the surrogate model and energy consumption of the two approaches. 



\section{Background}
\label{sec::background}

\subsection{Surrogate assisted models: Kriging and Kriging Partial Least Squares }

The aim of a surrogate model, also referred to as a meta-model, is to sufficiently approximate an estimate of the fitness values of a potential solution, while reducing the run time of the evolutionary process, compared to the run time of using the real fitness function alone. This requires that the surrogate model is well-posed and requires a robust model management strategy, otherwise, the evolutionary algorithm may converge to a false optimum~\cite{JIN201161}.



Kriging allows for interpolation by assuming spatial correlation exists between known data points. The process itself is informed by the distance and variation between these data points. More explicitly, a kernel function $K(\cdot)$ denotes the spatial correlation between two samples x$_i$ and x$_j$ as shown in Equation~\ref{eqn::kernel_dist},







\begin{equation}
K(x, x') = \prod_{i=1}^m exp(- \theta_i (x_i -x'_i)^2)
\label{eqn::kernel_dist}
\end{equation}
 
\noindent where the $\theta$ parameter determines the rate at which the correlation decays to zero and $m$ is the dimension. The $\theta$ parameter is determined by the Maximum Likelihood Estimator (MLE). One drawback, however, is that for large $m$, the cost increases significantly since the MLE algorithm requires calculating the inverse of the correlation matrix many times.


 Kriging Partial Least Squares (KPLS) seeks to address this by effectively reducing the number of parameters calculated~\cite{bouhlel2016improving}. It does so using partial least squares which project the high-dimensional data into a lower dimension using principal components. Equation \ref{eqn::kernel_dist2} details the KPLS kernel,

\begin{equation}
K(x, x') = \prod_{k=1}^{h} \prod_{i=1}^{m} exp(- \theta_k (w^{(k)}_{*i} x_i - w^{(k)}_{*i} x'_i)^2)
\label{eqn::kernel_dist2}
\end{equation}

\noindent where $w_*$ are rotated principal directions which maximize the covariance and are a measure of how important each principal component is. The number of principal components $h$ $<<$ $m$ which allows for the substantial improvement in computational cost associated with the KPLS approach. For full details of the method, please refer to~\cite{bouhlel2016improving}.

\subsection{Phenotypic Distance}

The rationale behind employing phenotypic distance draws inspiration not only from Stork et al.'s work~\cite{pheno_dist_kernel} but also from our research in Genetic Programming (GP)~\cite{9308386,galvan2022semantics}, focusing on semantics. Specifically, our contributions in the realm of semantic distance metrics include advancements in underexplored domains, such as multi-objective optimisation~\cite{DBLP:conf/gecco/GalvanS19,9308386,galvan2022semantics,stapleton2021semantic}. In traditional GP, semantics refer to the behaviour of a program given a finite set of inputs.

In the context of neuroevolution, we can define a solution sample $x$ as having the semantic or phenotypic behaviour of the $i^{th}$ program such that
$x_i = s(p_i)$, where the semantics $s(p)$ of a program $p$ is the vector of values from the output.  From Equation~\ref{eqn::kernel_dist} we can then define our phenotypic distance $d$ as

\begin{equation}
d(x_i, x_j) = d (s(p_i), s(p_j)))
\label{eqn::kernel_dist3}
\end{equation}

As such the distance metric is dependent on the outputs of each network. In this work, we use the convention established by Stork et al. ~\cite{pheno_dist_kernel}, where the $x_i$ is a flattened vector containing the output of the nodes at the final layer for all data instances. As such, our phenotypic distance vector length is given as the number of images of the validation dataset times the number of classes. It is important to note that this approach can be extended to any deep learning model architecture that can have its output represented in vectorised form, such as transformers, however, further research would be required to determine the limitations and scalability of applying this approach to other deep learning approaches.

\section{Methodology}
\label{sec::methodology}

\subsection{NeuroLGP}
\label{sub::neurolgp}



With the original LGP paradigm~\cite{brameier2007linear}, representation is inspired by the von Neumann architecture of modern CPUs which use registers for storing or modifying small amounts of data. The content of each register in this architecture can be changed using instruction operations. An instruction, in this context, has three main components: an \textit{operand}, \textit{registers} for which the operand operates on (one register for 2-register instruction and two registers for 3-register instruction) and the \textit{destination register} which stores the computed results. In the context of LGP, genetic operations work by either modifying these registers or the instructions that operate on them. While the original approach of LGP uses registers to store small amounts of data or instructions, it is possible to abstract this form so that the registers are themselves pointers to much larger amounts of data stored in memory. Our proposed approach does as such, using these pointers to instead control the flow of the initial and intermediary data from each outputted layer of our evolvable DNN. The left of Figure~\ref{fig:lgp_neuro_example}, demonstrates psuedocode how the expected representation would look, where $r[i]$ denotes the $i^{th}$ register. Each line of code is executed imperatively. This example contains both effective and non-effective lines of code. The non-effective lines of code are commented out and, as such, are not compiled. The register r[0] is a specially designated register for the final output of the program. To the best of our knowledge, LGP has not yet been used for neuroevolution.


\begin{figure}

\noindent\begin{minipage}{.47\textwidth}
\begin{lstlisting}[basicstyle=\scriptsize,,frame=tlrb]
def neuroLGP(...)
{ 
    r[0] := Conv(r[1])
    // r[4] := BatchNorm(r[3])
    r[5] := MaxPool(r[0])
    r[11] := BatchNorm(r[5])
    r[0] := Dense(r[11])   
    ...
}
\end{lstlisting}

\end{minipage}\hfill
\begin{minipage}{.47\textwidth}

\begin{lstlisting}[language=Python, basicstyle=\scriptsize,columns=fullflexible,frame=tlrb]
input = tf.keras.Input(shape=(imput_dim))

x = tf.keras.layers.Conv2D(32, 3, ...)(input)
# x = tf.keras.layers.BatchNormalization()(x)
x = tf.keras.layers.MaxPooling2D(3)(x)
x = tf.keras.layers.BatchNormalization()(x)
output = tf.keras.layers.Dense(x)

model = tf.keras.Model(output)

\end{lstlisting}

\end{minipage}
\caption{\textit{left:} NeuroLGP psuedocode for python. \textit{right:}  Functional API example in TensorFlow.
}
\label{fig:lgp_neuro_example}
\end{figure}

To understand how this representation can be useful for the case of neuroevolution, some example code in TensorFlow is given to demonstrate the imperative nature of defining models as seen in the right of Figure~\ref{fig:lgp_neuro_example}. This code demonstrates an example of a model definition using the functional API from TensorFlow. On Lines 3, 5 and 6, to the left of the assignment, a variable $x$ will hold the outputs of each statement and for Lines 5 -- 6, to the right of the assignment, $x$ is passed to each layer. As such, the variable $x$ holds transient data which updates as each line is executed. The aim is to replace $x$ using abstract or virtual registers (i.e r[0], r[1], r[2] ...etc), where the $x$ variables to the left of the assignment are represented using a \textit{destination register} and values to the right represent  \textit{registers} to be operated on. Line 4 is not executed.

\subsection{NeuroLGP with Surrogate Model (NeuroLGP-SM)}
\label{sub::neurolgp-sm}

\begin{figure}
 
  \centering
  \includegraphics[width=\textwidth]{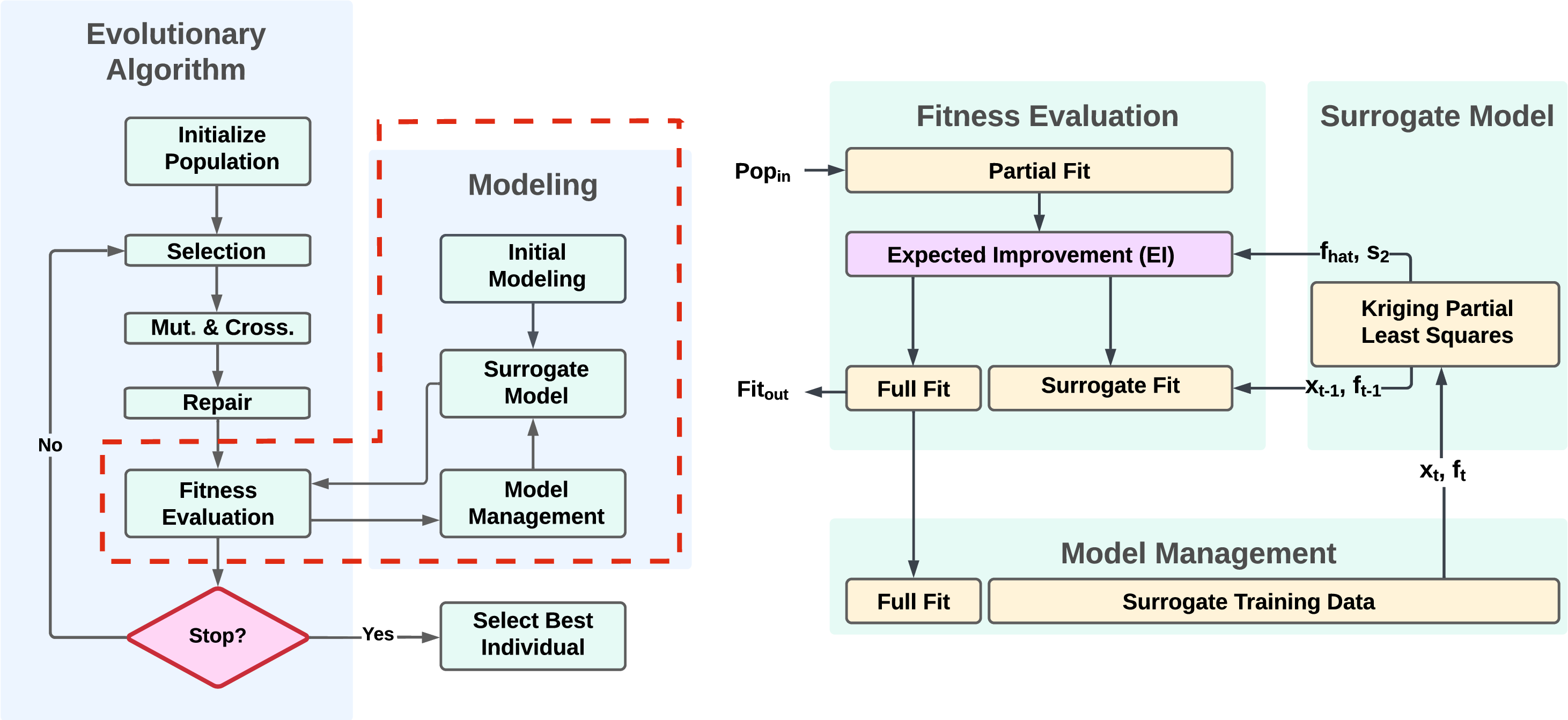}
  \caption{\textit{Left:} Diagram showing the interplay between a typical evolutionary algorithm and a surrogate model approach. \textit{Right:} The surrogate model management strategy is shown on a more granular level.}

  \label{fig:surrogate_overview}
\end{figure}

To determine which individuals should be fully evaluated, we use the expected improvement (EI) criteria~\cite{brochu2010tutorial,jones1998efficient}. EI is an acquisition function where the next candidate solution to be evaluated is based on the expected improvement over the current best solution so far. As such, EI allows us to evaluate solutions from regions of the search space likely to see the best improvement. To estimate the fitness using the surrogate model, first, we evaluate the fitness based on a limited number of epochs. From here, we split our population based on the EI criterion, whose parameters are informed by the surrogate model. Based on the EI criterion we fully evaluate a subset of the population, leaving the remaining proportion of the population to be evaluated using the surrogate model. Using the fully evaluated fitness we can now update the surrogate training data and re-train our surrogate model. In this sense, the use of the EI criterion is two-fold. Firstly, it allows us to evaluate individuals we expect to offer the greatest improvement, ensuring we expend our resources on the most promising networks. Secondly, it allows us to select iteratively, informing our surrogate model with better and better solutions. Figure~\ref{fig:surrogate_overview} demonstrates our framework. On the left, we can see the workflows of the evolutionary process and the surrogate model. On the right, we can see in more detail how the expected EI criterion is used within our surrogate model management strategy. 




\subsection{Genetic Operations and Repair Mechanism}
\label{sub::repair}

The mutation operator we use in this work mutates either a single input or output register or the operand. The mutation operator works on both effective code and non-effective code. Additionally, we make use of a novel effective crossover operator. This operator selects two crossover points from the effective sections of the parent code and then transfers segments to create offspring. While the ends of each segment contain effective code the code within may be non-effective. A further point mutation repair is applied at either segment end to ensure input and output registers match after crossover has been applied. Ultimately, this results in always producing a valid network.

A repair mechanism is incorporated when performing the \sloppy{genotype-to-phenotype} mapping to ensure models are compilable. There are several conditions for incorporating a repair mechanism, but it is important to note that in each case the repair is only performed on the effective code, by either inserting or deleting a specific layer.

\begin{itemize}
\item When there is no effective code in the genotype, we perform a single effective mutation inserting a single convolutional layer from our feature list. 

\item It is possible for the program to compile without a convolutional layer at the start (i.e., Max pooling, dropout, etc). We add an effective insertion of a convolutional layer before any layer that does not match this criteria.

\item When the input data dimensions are too low or if the number of data-reducing operations exceeds the dimensions of the data, the neural network will fail to compile, we remove layers iteratively until the model is valid.

\item If the input data dimensions are too high or if the number of data-reducing operations are too few, our fully-connected layer can be composed of an arbitrarily large number of nodes. Furthermore, adding further fully-connected layers can cause the models to become prohibitively expensive to evaluate, as such, we put additional conditions in to check and add a fully-connected layer before the final output layer. The size of the fully connected layer is dependent on the number of parameters from the CNN portion of the architecture. 

\end{itemize}

\section{Experimental Setup}
\label{sec::exp}


The Breast Cancer Histopathological Image Classification (BreakHis)~\cite{spanhol2016breast} dataset is a binary classification dataset consisting of microscopic images containing 2,480 benign and 5,429 malignant tumours, split across four types of magnification (40x, 100x, 200x and 400x). Each image consists of 700X460 pixels and 3-channel RGB with 8-bit depth in each channel. These images are re-scaled to a lower resolution of 64x64 pixels with 3-channels for each experiment. The overall data split for training, validation, test and test-2 is approximately (63.5\%, 12.5\%, 12.5\%, 12.5\%) or at the training level for each network, when only one test set is considered ~(70\%, 15\%, 15\%). The first test set is used to evaluate the accuracy of each network on unseen data and is used to generate a fitness function for the NeuroLGP approach. The second data set is used to evaluate the NeuroLGP approach after evolution and as such is unseen to both the neural network and evolutionary processes. Since the dataset is imbalanced, synthetic minority oversampling technique (SMOTE)~\cite{chawla2002smote} is used to up-sample the minority class. In our work we use image level accuracy (ILA), which is the standard classification accuracy at the image level~\cite{benhammou2020breakhis}.
We performed a substantial number of independent runs, 96 in all, totalling approximately 80 GPU days (3 approaches x 4 datasets x 8 runs). A generation size of 15 and a population size of 50 were selected for all 3 approaches. Experiments were conducted using Kay supercomputer provided by the Irish Centre for High-End Computing (ICHEC). Runs were run in parallel, with each individual run assigned to a single Nvidia Tesla V100 GPU with 16GB Ram. A set of three experiments were designed to test three approaches: 
\begin{enumerate}
\item A \textit{\Baseline{}} approach which is a random search approach that initializes the network architecture layers randomly. These networks are not evolved but are repaired in line with the repair mechanism as discussed in Section~\ref{sub::repair}. All networks are fully trained to the full number of epochs (30 epochs), 
\item The \textit{\Expensive{}} approach where we evolve the structure of our networks using the NeuroLGP approach for the max number of epochs (Pop. size = 50, Gen. size =15), 
\item The \textit{\Surrogate{}} approach where we employ the NeuroLGP-SM. The surrogate model is informed using partially trained networks (10 epochs) for 60\% of the population size.
\end{enumerate}

\section{Results}
\label{sec::results}

\subsection{Preliminary Analysis of the \BaselineCap{} Model}

\begin{figure*}[h]
  \centering
  
  \includegraphics[width=0.68\linewidth]{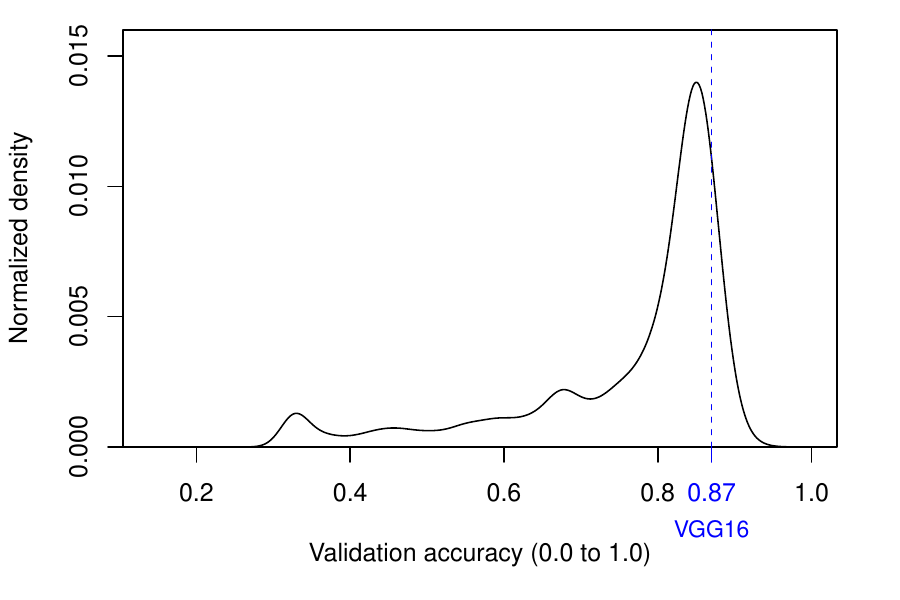}
  \caption{Accuracy of 6000 individuals (read text).}

  \label{fig:density}
\end{figure*}

The \Baseline{} in our approach makes use of random search over the feature extraction portion of our network architecture with a set of repair operations to ensure the architecture can be compiled. Typical network sizes from this approach ranged from $\sim$20K to $\sim$12.5M parameters with the \Baseline. For example of VGG-16 has $\sim$40M parameters~\cite{simonyan2014very}. Figure~\ref{fig:density} shows the distribution of accuracy values for the \Baseline, for magnification x200, for all networks across 8 runs (750 individuals per run, 6000 individuals overall). It is important to point out that across all 8 runs, every model found by the \Baseline{} was valid and compilable. 
Slight peaks at 0.33 and 0.66 represent poor performing networks that that classify all the data as a single class. We can see that despite a tailed distribution to the left, the vast majority of individuals are centred around the median peak of 0.83. Additionally, the performance of VGG-16 is also shown in the figure with an accuracy of 0.87, which was verified in our own experiments and documented in the work by Cascianelli et al.~\cite{cascianelli2018dimensionality}. The main takeaway is the \Baseline{} approach can find competitive architectures to compare against the \Surrogate{} and \Expensive{} approaches.

\subsection{Comparison of Models}

\newcommand{\maxRanXforty}{0.889}
\newcommand{\maxRanXhundred}{0.869}
\newcommand{\maxRanXtwohundred}{0.946}
\newcommand{\maxRanXfourhundred}{0.914}

\newcommand{\maxSurXforty}{0.913}
\newcommand{\maxSurXhundred}{0.903}
\newcommand{\maxSurXtwohundred}{0.970}
\newcommand{\maxSurXfourhundred}{0.925}

\newcommand{\maxExpXforty}{0.930}
\newcommand{\maxExpXhundred}{0.916}
\newcommand{\maxExpXtwohundred}{0.960}
\newcommand{\maxExpXfourhundred}{0.925}

A recent survey by Benhammou on the BreakHis dataset lists a comprehensive comparison of various machine learning approaches in terms of accuracy~\cite{benhammou2020breakhis}. The results demonstrated here match the state-of-the-art approaches in terms of performance and are particularly notable as the approach we use does not make use of transfer learning techniques.

\begin{figure*}[h]
  \centering
  \begin{subfigure}{0.45\linewidth}
    \includegraphics[width=\linewidth]{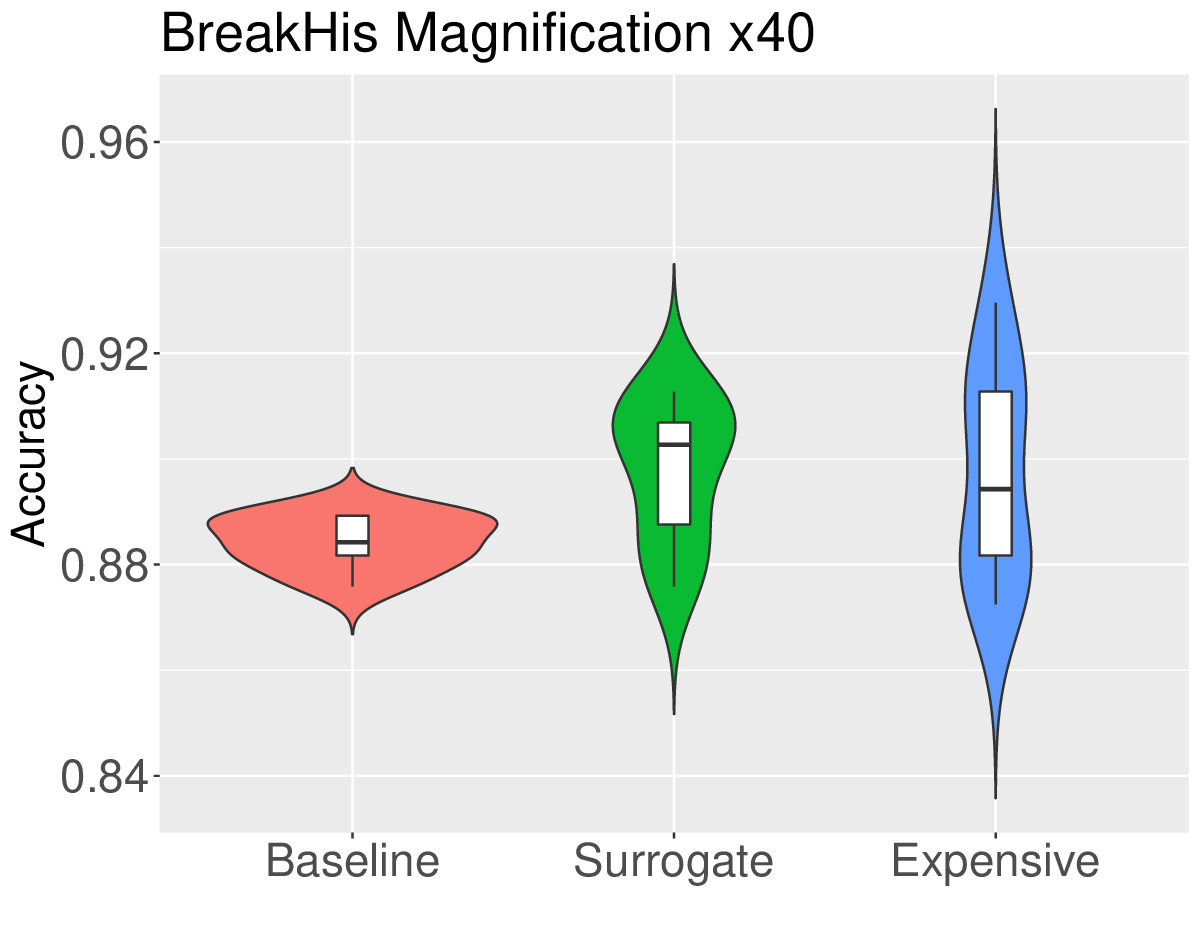}
    \caption{}
    \label{fig:violin40}
  \end{subfigure}
  \hfill
  \begin{subfigure}{0.45\linewidth}
    \includegraphics[width=\linewidth]{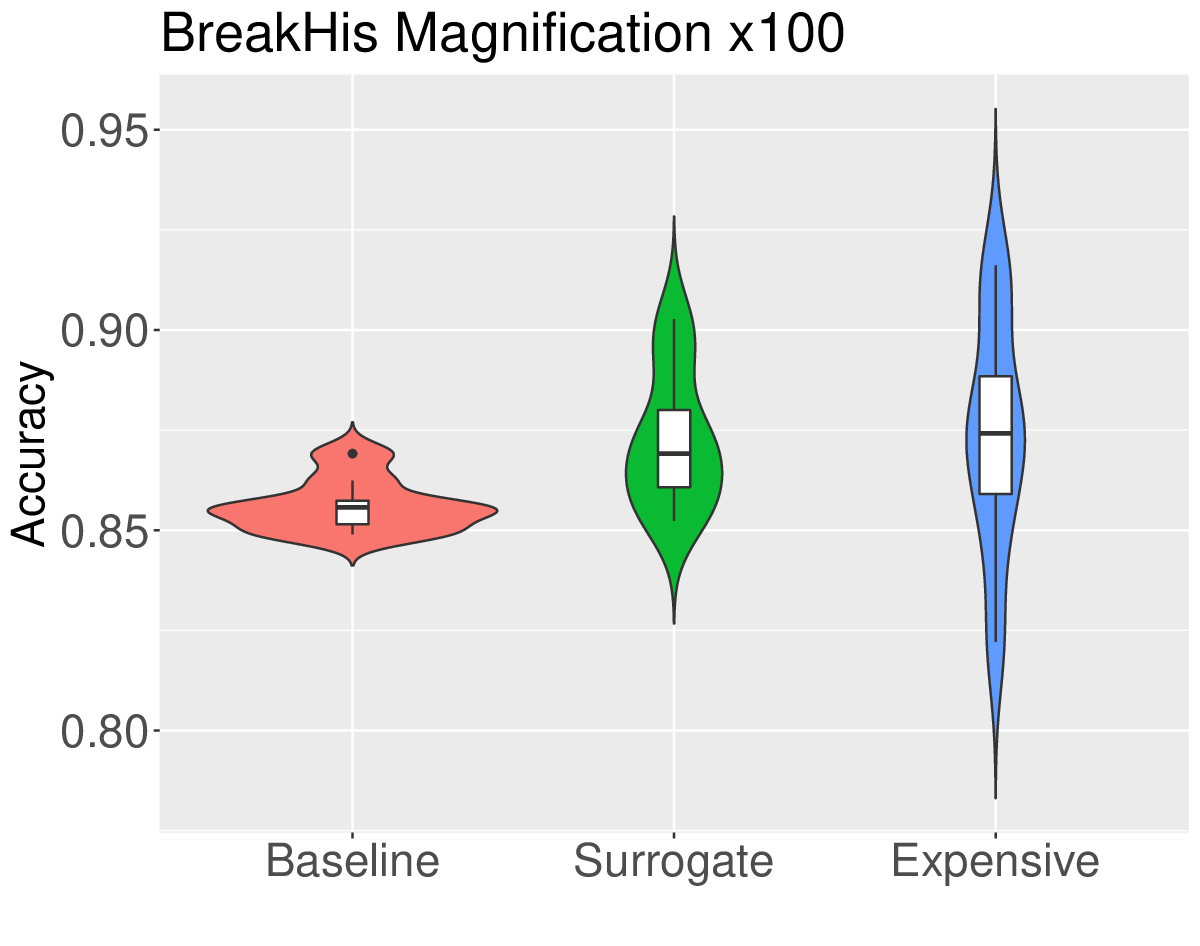}
    \caption{}
    \label{fig:violin100}
  \end{subfigure}

  \begin{subfigure}{0.45\linewidth}
    \includegraphics[width=\linewidth]{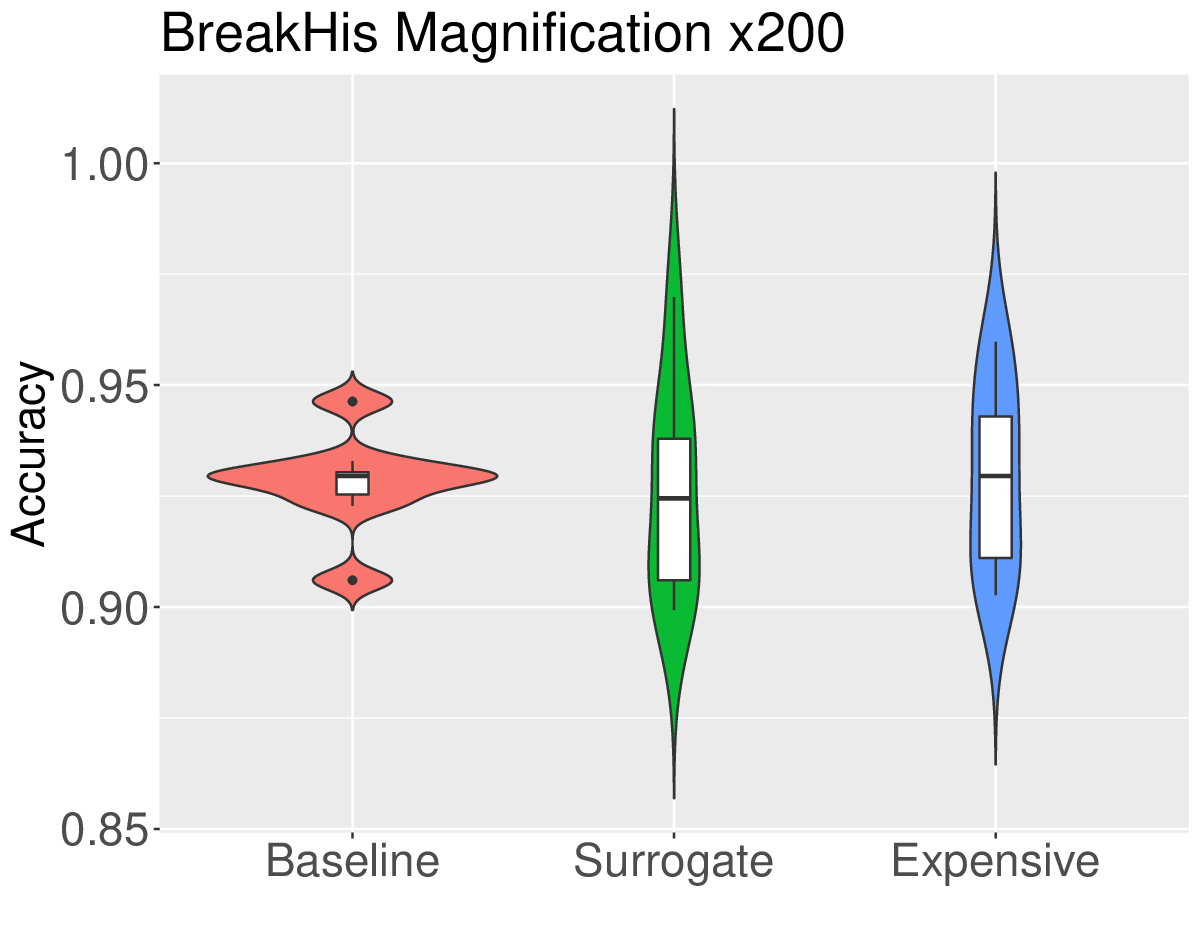}
    \caption{}
    \label{fig:violin200}
  \end{subfigure}
  \hfill
  \begin{subfigure}{0.45\linewidth}
    \includegraphics[width=\linewidth]{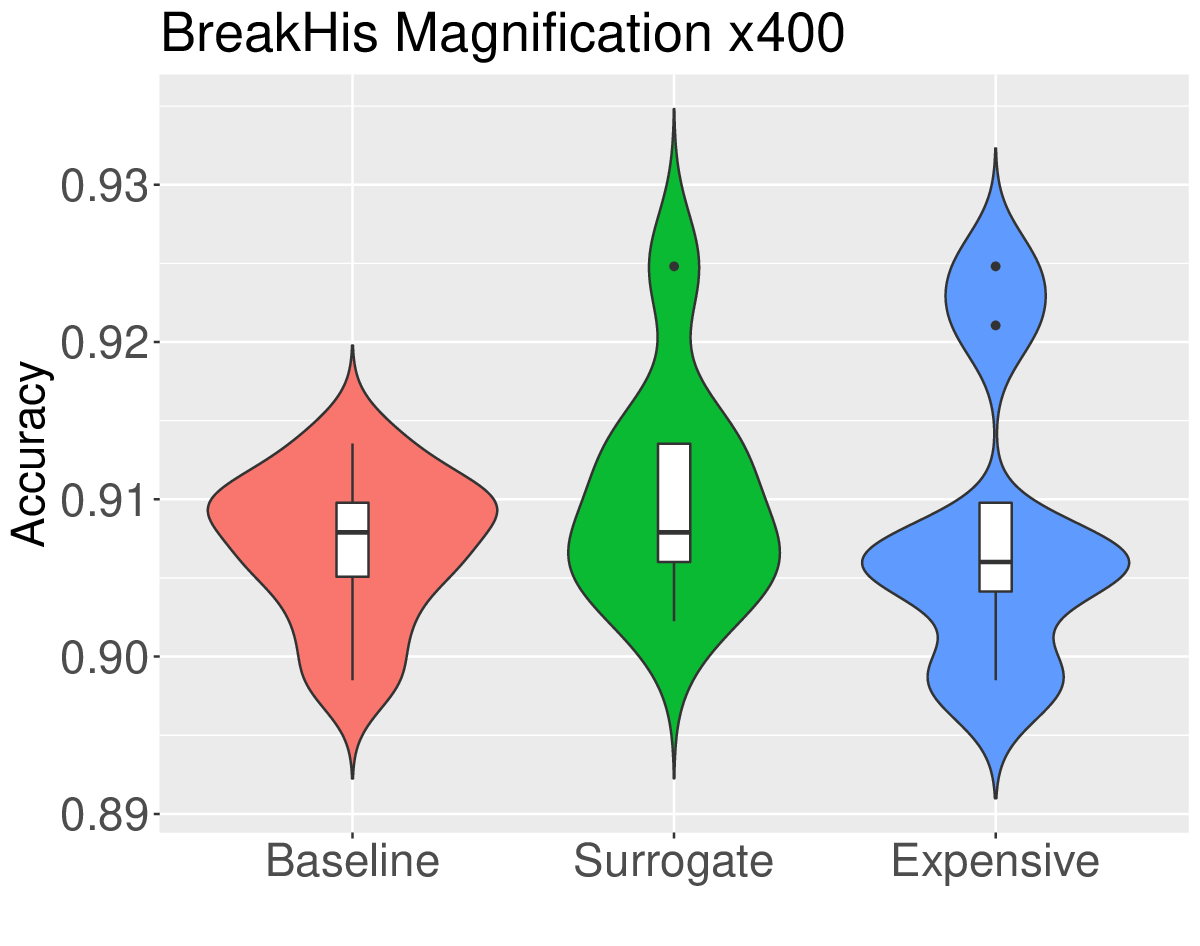}
    \caption{}
    \label{fig:violin400}
  \end{subfigure}
  \caption{Figures~\ref{fig:violin40} through~\ref{fig:violin400} plot violin plots, showing the distribution of accuracies of the architectures with the best fitness for each of the four data sets across 8 runs.} 
  \label{fig:2x2formation}
\end{figure*}

A comparison is conducted between the three experiments as outlined in Section~\ref{sec::exp} for the BreakHis dataset. The violin plots in Figures~\ref{fig:2x2formation}(a-d) plot the accuracies of the best individual in terms of fitness, for each run, for the x40, x100, x200 and x400 magnifications, respectively. For reference, if we were to look at \Baseline{} as an example (left-hand side in each plot of Figure~\ref{fig:2x2formation}), then the individuals shown in these plots represent individuals found at the far right of the density plot in Figure~\ref{fig:density}. Initially, we ran the three models for 4 runs only, which took approximately 40 GPU days in total. It should be noted that in neuroevolution the norm is usually to do a single run. The initial results showed a tendency for the \Surrogate{}  and \Expensive{}  models to outperform the \Baseline{} model for the x40, x100 and x400 models however for x200 the \Surrogate{} and \Expensive{} underperformed. It was decided to extend this, to 8 runs totalling 80 GPU days in all. The updated results are represented here in the violin plots, where again we can see that in the case of x40 and x100 magnification, the \Surrogate{} (green) and \Expensive{} (blue) models tend to outperform the \Baseline{} (red) model. Results are more mixed when we look at x200, but is markedly improved over the initial set of 4 runs. The x400 magnification appears somewhat mixed, however, with more runs x400 may have more separation for \Surrogate{} and \Expensive{} models over the \Baseline{} given there are few better-performing individuals above the 92\% mark. Comparing \Surrogate{} and \Expensive{} models in particular, an important observation can be made, in that for all magnifications they have similar accuracies. Given the distribution of the baseline is not fully separable from the other methods, compounded by the low run number, it was decided analysis of the best individuals across all runs would be more informative rather than relying purely on statistical testing.



The accuracies for the best-performing individuals across all runs for the \Baseline{}, \Surrogate{} and \Expensive{} models are listed respectively as such; for the x40 magnification the accuracies are 
\maxRanXforty{}, \maxSurXforty {} 
 and \maxExpXforty{}; for the x100 magnification are 
\maxRanXhundred{}, \maxSurXhundred {} 
 and \maxExpXhundred{}; for the x200 magnification are 
\maxRanXtwohundred{}, \maxSurXtwohundred {} 
 and \maxExpXtwohundred{} and finally for the x400 magnification are 
\maxRanXfourhundred{}, \maxSurXfourhundred {} 
 and \maxExpXfourhundred{}. In no instance did the \Baseline{} produce the best-performing individual. For the x40 and x100 magnifications, the \Expensive{} model produced the best-performing individual. For the x200 magnification the \Surrogate{} model produced the best-performing individual and the x400 magnification has a tie for the best-performing individual. In summary, while on average our \Surrogate{} and \Expensive{} models are as good as or better than the \Baseline{}, when we consider the best-performing individuals across all runs the \Surrogate{} and \Expensive{} models are better.


\subsection{Analysis of the \SurrogateCap{} Model}

\newcommand{\mseXforty}{0.0037}
\newcommand{\mseXhundred}{0.0017}
\newcommand{\mseXtwohundred}{0.0014}
\newcommand{\mseXfourhundred}{0.0009}

\newcommand{\kendallXforty}{0.6019}
\newcommand{\kendallXhundred}{0.6791}
\newcommand{\kendallXtwohundred}{0.6225}
\newcommand{\kendallXfourhundred}{0.5647}

\newcommand{\codXforty}{0.5026}
\newcommand{\codXhundred}{0.6665}
\newcommand{\codXtwohundred}{0.7079}
\newcommand{\codXfourhundred}{0.7786}

\newcommand{\timeSurXforty}{15.9}
\newcommand{\timeSurXhundred}{16.6}
\newcommand{\timeSurXtwohundred}{16.8}
\newcommand{\timeSurXfourhundred}{15.3}

\newcommand{\timeSurStdXforty}{0.7}
\newcommand{\timeSurStdXhundred}{0.7}
\newcommand{\timeSurStdXtwohundred}{0.8}
\newcommand{\timeSurStdXfourhundred}{0.6}

\newcommand{\timeExpXforty}{21.3}
\newcommand{\timeExpXhundred}{22.7}
\newcommand{\timeExpXtwohundred}{22.4}
\newcommand{\timeExpXfourhundred}{20.0}

\newcommand{\timeExpStdXforty}{0.2}
\newcommand{\timeExpStdXhundred}{0.1}
\newcommand{\timeExpStdXtwohundred}{0.1}
\newcommand{\timeExpStdXfourhundred}{0.1}

\newcommand{\timePercXforty}{25.3}
\newcommand{\timePercXhundred}{26.7}
\newcommand{\timePercXtwohundred}{24.9}
\newcommand{\timePercXfourhundred}{23.8}

We use three metrics to determine how well our surrogate model performs in terms of predicting
the fitness of our partially trained models. Firstly, we use the Mean Squared Error (MSE) between the predicted fitness and actual fitness. Values closer to 0 indicate our surrogate model is accurate in predicting fitness. Secondly, Kendall's Tau is used to measure the correlation between the predicted fitness and the actual fitness~\cite{sun2019evolving}. A coefficient value of -1 indicates a perfect negative correlation, while a coefficient of +1 indicates a perfect positive correlation. Values close to 0 would indicate no discernible correlation. When considering all runs, a mean value close to 0 would indicate our surrogate model is performing poorly while a mean value closer to 1 would indicate an ideal-performing surrogate model for Kendall's Tau. Thirdly, the R$^2$ score~\cite{bhosekar2018advances}, is used to measure how close the predicted value is to its true value and ranges from -$\infty$ to 1, and explains how much variability is in the prediction model. R$^2$ scores close to 1 would represent a model which perfectly fits the data. 

\begin{table}[h]
  \centering
  \caption{Average quality of fit and number of GPU hours per run.}
  \begin{tabular}{|l||c|c|c||cc|cc|c|}
    \hline
     & \multicolumn{3}{c||}{\textbf{ Quality of fit}} & \multicolumn{5}{c|}{\textbf{GPU hours per run}} \\
    & \multicolumn{3}{c||}{\textbf{ NeuroLGP-SM}} & \multicolumn{5}{c|}{\textbf{}} \\
    \hline
    \textbf{Mag.} & \textbf{MSE} & \textbf{Kendall's} & \textbf{R$^2$} & \multicolumn{2}{c|}{\textbf{Expensive (hrs)}} & \multicolumn{2}{c|}{\textbf{Surrogate (hrs)}} & \textbf{Reduction } \\
    & & \textbf{Tau} & & \textbf{Mean} & \textbf{Std} & \textbf{Mean} & \textbf{Std} & \textbf{Time} \\
    \hline
    x40 & \mseXforty & \kendallXforty & \codXforty & \timeExpXforty & $\pm$ \timeExpStdXforty & \timeSurXforty & $\pm$ \timeSurStdXforty & \timePercXforty $\%$ \\
    x100 & \mseXhundred & \kendallXhundred & \codXhundred & \timeExpXhundred & $\pm$ \timeExpStdXhundred & \timeSurXhundred & $\pm$ \timeSurStdXhundred & \timePercXhundred $\%$ \\
    x200 & \mseXtwohundred & \kendallXtwohundred & \codXtwohundred & \timeExpXtwohundred & $\pm$ \timeExpStdXtwohundred & \timeSurXtwohundred & $\pm$ \timeSurStdXtwohundred & \timePercXtwohundred $\%$ \\
    x400 & \mseXfourhundred & \kendallXfourhundred & \codXfourhundred & \timeExpXfourhundred & $\pm$ \timeExpStdXfourhundred & \timeSurXfourhundred & $\pm$ \timeSurStdXfourhundred & \timePercXfourhundred $\%$ \\
    \hline
  \end{tabular}
  \label{tab:fit}
\end{table}

Table~\ref{tab:fit} summarizes the effectiveness of the \Surrogate{} model for each of the datasets, as seen in the `Quality of fit NeuroLGP-SM' header of this table. The low MSE values (second column), show a relatively low error between the predicted and actual fitness, however, the MSE value for x40 magnification was comparatively high at \mseXforty, compared to x100, x200 and x400 at magnifications \mseXhundred{}, \mseXtwohundred{} and \mseXfourhundred{} respectively. Furthermore, Kendall's Tau values range from \kendallXfourhundred{} to \kendallXhundred{} across the four datasets indicating a strong positive correlation between the actual and predicted fitness. We can see that for x40 we have R$^2$ of \codXforty{} indicating a moderate level of fit being captured by the surrogate model while x100, x200 and x400 we have R$^2$ values of \codXhundred{}, \codXtwohundred{} and \codXfourhundred{} indicating a strong level of fit. A deeper dive into why x40 had a comparatively lower performance revealed that for a particular run, the surrogate model estimated poorly on the lower fitness models but in general was better for higher fitness models and was better overall for the other seven runs.

The `GPU hours per run' header of Table~\ref{tab:fit} shows a comparison between the \Expensive{} and \Surrogate{} model in average runtime per GPU hour, in the fifth and sixth columns respectively.  Of note is that the \Surrogate{} model typically is an order of magnitude higher for standard deviation, meaning there is greater variance associated with runtime for the \Surrogate{} model approach. The last column shows the percentage in terms of time saved using the \Surrogate{} model over the \Expensive{} model. In general, we can see that given the parameters selected, we roughly save 25\% in terms of GPU hours. Overall, if we consider the total time it took to run 8 runs for all  4 datasets, the \Expensive{} models took $\sim$28 GPU days and the \Surrogate{} models $\sim$21 GPU days, saving approximately $\sim$7 GPU days in all.

\section{Conclusion and Future Work}
\label{sec::conclusion}


In our work, we adapt the use of phenotypic distance vectors for surrogate-assisted neuroevolution of Deep Neural Networks (DNNs) using a variation of Kriging, entitled Kriging Partial Least Squares (KPLS). While phenotypic distance vectors have previously been used for traditional Artificial Neural Networks (ANNs) of only a few layers, their use in surrogate modelling for DNNs marks a significant leap forward in tackling high-dimensionality in neuroevolution. Our surrogate model management strategy makes use of a novel neuroevolutionary approach inspired by Linear Genetic Programming, entitled NeuroLGP which allows us to evolve compact, robust and variable-length architectures. This new approach was shown to outperform a competitive baseline for both the expensive (NeuroLGP) and surrogate-based variant
(NeuroLGP-SM) using four magnification subsets of the BreakHis dataset when considering best-individuals across all runs. While we conducted an extensive set of experiments totalling 80 GPU days, in future, we would like to extend the number of runs from 8 to 16 to conduct a full statistical analysis. Furthermore, the runtime associated with NeuroLGP-SM was on average 25\% faster compared to the expensive variant. 

\section*{Acknowledgements}

This publication has emanated from research conducted with the financial support of Science Foundation Ireland under Grant number 18/CRT/6049. For the purpose of Open Access, the author has applied a CC BY public copyright licence to any Author Accepted Manuscript version arising from this submission.

\bibliographystyle{abbrv}
\bibliography{ref.bib}

\end{document}